\documentclass[11pt,a4paper]{article}
\pdfoutput=1
\usepackage[T1]{fontenc}
\usepackage{float}
\usepackage{indentfirst} 
\usepackage{amsmath} 
\usepackage{amsfonts} 
\usepackage[authoryear, sort&compress]{natbib} 
\usepackage{hyperref} 
\hypersetup{colorlinks=true, citecolor=blue}
\usepackage{graphicx}
\usepackage{subcaption}
\usepackage{booktabs} 
\usepackage{multirow} 
\usepackage{siunitx} 
\usepackage{caption}  
\usepackage{newtxtext, newtxmath} 
\usepackage[normalem]{ulem}
 
\title{FADConv: A Frequency-Aware Dynamic Convolution for Farmland Non-agriculturalization Identification and Segmentation}
\date{} 
\author{Tan Shu , Li Shen*}
\begin{document}
	\maketitle
	\section*{Abstract}
	Cropland non-agriculturalization refers to the conversion of arable land into non-agricultural uses such as forests, residential areas, and construction sites. This phenomenon not only directly leads to the loss of cropland resources but also poses systemic threats to food security and agricultural sustainability. Accurate identification of cropland and non-cropland areas is crucial for detecting and addressing this issue. Although remote sensing and deep learning methods have shown promise in cropland segmentation, challenges persist in misidentification and omission errors, particularly with high-resolution remote sensing imagery.Traditional CNNs employ static convolution layers, while dynamic convolution studies demonstrate that adaptively weighting multiple convolutional kernels through attention mechanisms can enhance accuracy. However, existing dynamic convolution methods relying on Global Average Pooling (GAP) for attention weight allocation suffer from information loss, limiting segmentation precision. This paper proposes Frequency-Aware Dynamic Convolution (FADConv) and a Frequency Attention (FAT) module to address these limitations. Building upon the foundational structure of dynamic convolution, we designed FADConv by integrating 2D Discrete Cosine Transform (2D DCT) to capture frequency domain features and fuse them. FAT module generates high-quality attention weights that replace the traditional GAP method,making the combination between dynamic convolution kernels more reasonable.Experiments on the GID and Hi-CNA datasets demonstrate that FADConv significantly improves segmentation accuracy with minimal computational overhead. For instance, ResNet18 with FADConv achieves 1.9\% and 2.7\% increases in F1-score and IoU for cropland segmentation on GID, with only 58.87M additional MAdds. Compared to other dynamic convolution approaches, FADConv exhibits superior performance in cropland segmentation tasks.
	
	\vspace{0.5cm} 
	\noindent\rule{\textwidth}{0.4pt} 
	\noindent \textbf{Author Contributions:} This work was researched and written by Tan Shu. Li Shen led the project and provided academic guidance.\\
	\textsuperscript{*}Corresponding author. Email: ergoproxy@my.swjtu.edu.cn.
	\section{Introduction}
	With the acceleration of urbanization in human society, the area of cropland are gradually shrinking. The conversion of cropland to construction land causes non-agriculturalization of cropland, often leading to irreversible damage to soil structure and posing significant threats to agricultural resources and ecosystems. Due to the impact of illegal construction and afforestation, the safety of grain production has been threatened.  In response to this phenomenon,Chinese government propose the concept of "Cropland Non-agriculturalization" (CNA) to address this issue \citep{sun2024}.
	
	In recent years, remote sensing technology has gradually replaced traditional manual surveys due to its advantages of wide monitoring coverage, rapid data acquisition, and rich information capture \cite{Wang2024}. Compared to low-resolution imagery, high-resolution images can reveal previously undetectable targets and display more detailed textures, shapes, and spatial patterns of features \citep{Kemker2018}. Deep learning and machine learning have achieved remarkable progress in computer vision, offering robust capabilities for large-scale data processing and feature extraction. Combining high-resolution remote sensing imagery with semantic segmentation techniques has become an effective approach for addressing non-agriculturalization. However, challenges remain in accurate pixel-level cropland segmentation due to crop diversity, seasonal variations, and ambiguous boundaries.
	\\
	\\
	\textbf{Related Work and Limitations:}Prior to the rise of deep learning, cropland segmentation relied on traditional machine learning methods, ranging from clustering algorithms to Support Vector Machines (SVM) \citep{Cortes1995}, which required extensive manual feature engineering for pixel-level classification. The emergence of AlexNet \citep{Krizhevsky2012} demonstrated the potential of deep learning in image recognition, laying the foundation for subsequent advancements. ResNet \citep{He2016} introduced residual connections, proving that deeper networks enhance feature extraction. U-Net \citep{Ronneberger2015}, initially designed for medical image segmentation, achieved outstanding performance in other domains with its encoder-decoder architecture and multi-scale skip connections. The Fully Convolutional Network (FCN) \citep{Shelhamer2017} enabled end-to-end semantic segmentation through deconvolutional structures. The DeepLab series \citep{Chen2015,chen2017,chen2018,chen2018_2} pioneered atrous convolution and the Atrous Spatial Pyramid Pooling (ASPP) module to capture multi-scale contextual information. MobileNet \citep{Howard2017} reduced computational costs through depthwise separable convolutions, while subsequent versions improved efficiency with inverted residuals and linear bottlenecks.Squeeze-and-Excitation Networks (SENet) \citep{Hu2018} introduced the Squeeze-and-Excitation (SE) module as channel attention to dynamically adjust feature map weights. FCANet \citep{Qin2021} redefined channel attention by incorporating 2D Discrete Cosine Transform (DCT) for frequency-aware feature fusion. OrthoNets \citep{Salman2023} leveraged the orthogonality of DCT to enhance attention mechanisms. Vision Transformers (ViT)\citep{Dosovitskiy2021} demonstrated the potential of self-attention \citep{Vaswani2017} in capturing long-range dependencies but required substantial computational resources.Dynamic convolution methods, such as CondConv \citep{Yang2019} and DYConv \citep{chen_y2020}, improved model adaptability by dynamically weighting multiple expert kernels. ODConv \citep{Li2022} extended attention to input/output channels and kernel parameters.KernelWarehouse \citep{Li2024} reduces the number of parameters and enhances parameter reusability through kernel partitioning and warehouse sharing, achieving excellent results.The semantic segmentation approach based on deep learning and the dynamic convolution methods have great potential for addressing CNA issues and demonstrates high efficiency and practicality in the task of identifying and extracting cultivated and non-cultivated land areas.However, there are still some limitations:
	
	(1)Static convolution kernels in conventional CNNs exhibit limited receptive fields, constraining their ability to capture global contextual information. Their focus on local features often leads to misclassification of large homogeneous regions with similar spectral characteristics.
	
	(2)While current approaches to enhance feature extraction capacity primarily involve deepening network layers or increasing channel width, this strategy incurs a significant computational overhead – often exponentially proportional to architectural expansions.
	
	(3)Existing dynamic convolution methods enhance model performance by dynamically adjusting the weights of multiple parallel kernels using input features and global contextual information. However, due to the heavy reliance on attention mechanisms, these approaches typically employ Global Average Pooling (GAP) for global feature extraction, which treats the importance of all spatial positions as uniform. This simplistic aggregation fails to adequately capture the intricate spatial dependencies and frequency characteristics inherent in high-resolution imagery, resulting in feature information loss. Consequently, this limitation remains a bottleneck hindering further improvements in dynamic convolution methods.
	\\
	\\
	\textbf{Our Contributions:}To address these challenges, we propose Frequency-Aware Dynamic Convolution (FADConv), a novel dynamic convolution method that adapts kernel parameters based on frequency domain characteristics of the input, thereby enhancing network representational capacity. FADConv leverages 2D DCT to extract frequency features, improving upon conventional attention mechanisms. Specifically, we design the Frequency Attention (FAT) module, which derives frequency-aware attention weights by transforming spatial features into the frequency domain via 2D DCT and fusing multi-spectral components.The proposed FADConv dynamically adjusts expert kernel weights using frequency-guided attention, enabling convolutional operations to adapt to input features. By integrating the advantages of dynamic convolution with frequency domain features, FADConv expands model capacity with minimal computational overhead while achieving significant performance gains. At the same time, it integrates the frequency information extracted from the FAT module to dynamically adjust the weights of the expert convolution kernels for feature extraction, which substantially improves segmentation accuracy in diverse complex scenarios (e.g., ambiguous boundaries between cropland and non-agricultural areas).
	
	Furthermore, we generalize the FAT module to existing dynamic convolution frameworks (e.g.,ODConv \citep{Li2022},KW\citep{Li2024}), achieving good improvement effects compared with the original methods, validating the versatility of frequency-aware attention in enhancing feature discrimination and generalization.
	
	\section{Related Work}
	\noindent\textbf{Channel Attention:}We review the evolution of channel attention mechanisms. SENet \citep{Wang2020} pioneered channel attention through its Squeeze-and-Excitation (SE) moduleIt uses Global Average Pooling (GAP) to compress global information and then employs fully connected layers to learn the relationships between channels, selectively enhancing or suppressing channels. GSoP-Net \citep{Gao2019} improved channel attention by modeling higher-order channel interactions using second-order covariance. SRM \citep{Lee2019} integrated style transfer principles, enhancing global feature aggregation through mean and standard deviation pooling. ECANet \cite{Wang2020} replaced the SE module’s fully connected layers with 1D convolutions to achieve local cross-channel interactions. FCANet \citep{Qin2021} introduced frequency domain analysis into channel attention, employing Discrete Cosine Transform (DCT) to extract frequency-specific features for channel fusion. OrthoNets \citep{Salman2023} demonstrated that the inherent orthogonality in DCT is the key to modeling efficient channel attention, and propose randomized orthogonal filters as an alternative to GAP and DCT.
	\\
	\\
	\textbf{Dynamic Convolution:}Dynamic convolution methods enhance network adaptability by adjusting convolution weights based on input characteristics rather than relying solely on static training parameters. CondConv \citep{Yang2019} and DYConv \citep{chen_y2020} pioneered this paradigm by generating multiple parallel expert kernels and dynamically aggregating them using attention weights derived from GAP-processed features. ODConv \citep{Li2022} extended dynamic convolution mechanisms to four dimensions (inchannel,outchannel, kernel, and expert), enabling full-dimensional dynamic modulation. KernelWarehouse (KW) \citep{Li2024} optimized parameter efficiency through kernel partitioning and cross-layer sharing, achieving notable improvements on ImageNet and COCO datasets.
	
	Building upon these foundations, we introduce frequency domain features into dynamic convolution’s attention mechanism. Specifically, we replace traditional GAP with a frequency-aware feature extraction method that applies 2D DCT to each channel, capturing multi-frequency characteristics. These frequency features are then processed through fully connected layers to generate dynamic attention weights for expert kernel aggregation. The subsequent section details the architecture of FADConv.
	
	\section{Method}
	In this section, we introduce the overall architecture of FADConv . We then present the implementation details of its core module, the Frequency Attention(FAT) mechanism.
	\subsection{Review and Correlation of Dynamic Convolution}
	The dynamic convolutional layer generates the final convolutional kernel by linearly combining N parallel expert kernels, where the attention weights for kernels are dynamically conditioned on the input features to compute weighted aggregation of expert kernels.A fundamental formula can express this:
	\begin{equation}
		Y=(\alpha_1W_1+\alpha_2W_2+...+\alpha_eW_e)\ast X
	\end{equation}
	The dynamic attention mechanism adjusts the value of a based on the input to dynamically weight the expert convolution kernels.A critical challenge in dynamic convolution lies in efficiently deriving attention weights for expert kernels, as these weights are dynamically determined by input feature interactions.To address this, the proposed FADConv employs a three-stage dynamic mechanism for feature refinement:
	
	(1) Feature Extraction: Multiple Frequency features are extracted using 2D Discrete Cosine Transform(2D DCT).
	
	(2) Frequency Attention(FAT): A learnable frequency-domain fusion network dynamically allocates kernel weights.
	
	(3) Dynamic Kernel Generation: A parallel expert kernel architecture generates the final convolutional kernel.
	
	By transforming input features from the spatial to the frequency domain via 2D DCT which exhibits superior energy compaction properties,FADConv efficiently aggregates channel-wise,
	spatial, and frequency-domain features. The FAT module propagates global frequency-domain information to expert kernels, enabling optimal dynamic weight allocation.
	\subsection{Frequency-Domain Dynamic Convolution}
\begin{figure}[htbp]
		\centering
		\includegraphics[width=0.5\textwidth]{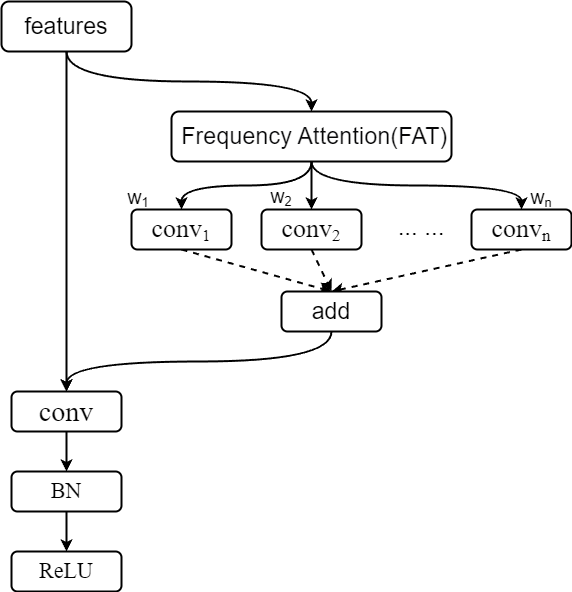}
		\caption{Overall structure of the Frequency-Aware Dynamic Convolution layer}   
		\label{fig:3-1} 
	\end{figure}

	Here, we present a frequency-domain enhanced dynamic convolution framework. Similar to DYConv\citep {chen_y2020} ,FADConv maintains K expert kernels alongside static convolutional parameters, allowing direct replacement of standard convolution layers. The FDconv design is shown in Figure \ref{fig:3-1}.The final kernel weights are aggregated via FAT, followed by Batch Normalization(BN) and ReLU activation to form a complete Frequency-Aware Dynamic Convolution Layer.The FADConv operation is defined as:
	\begin{equation}
		W_{\text{final}} = \sum_{e=1}^{K} \alpha_e(f) W_e
	\end{equation}
	\begin{equation}
		b_{\text{final}} = \sum_{e=1}^{K} \alpha_e(f) b_e
	\end{equation}
	\begin{equation}
		Y = W_{\text{final}}\ast X + b_{\text{final}}
	\end{equation}
	
	$K$ is the number of expert kernels, $\alpha_e(f)$ denotes frequency-aware attention weights,and represents frequency-related operations. $W_{\text{final}}$ and $b_{\text{final}}$ are linear combinations of expert kernels. The frequency variables $\alpha_e(f)$ obtained through 2D DCT are treated as nonlinear functions of the input features, allowing FDConv to inherently extract frequency information before convolution. This design enables FADConv to act as a prior-enhanced convolution, where attention mechanisms "reveal" critical frequency patterns to expert kernels during both forward and backward propagation. By integrating energy-compact frequency components, FADConv achieves comprehensive feature fusion.
	
	Secondly, we consider the computational cost from the perspectives of parameters and MAdds.The 2D DCT operation introduces no additional parameters, and the frequency fusion module incurs negligible parameter overhead, making FADConv’s parameter count comparable to DYConv. Compared to standard convolution, the extra Multiply-Adds (MAdds) introduced by FADConv are calculated as:
	\begin{align}
		MAdds_{\text{extra}} &= 
		MAdds_{\text{DCT}} + 
		MAdds_{\text{FAT}} +
		MAdds_{\text{dynamic\_kernel}}\nonumber\\
		&= 2C_{\text{in}} P^3 + \frac{C^2_\text{in}+C_\text{in}n}{r}
		+\frac{n(C_\text{in}C_\text{out}k^2)}{groups}+b
		\label{MAdds}
	\end{align}
	
	where $P$ is the DCT pooling size, $r$ is the channel reduction ratio, $n$ is the number of expert kernels, and $b$ represents negligible operations (e.g., pooling). The additional MAdds remain minimal compared to standard convolution
	$\frac{hwC_\text{in}C_\text{out}k^2}{groups}$.
	
	In addition,FADConv is architecture-agnostic and compatible with diverse CNN backbones. It supports arbitrary convolutional configurations (e.g., kernel size, stride, dilation, groups) without constraints, ensuring excellent portability and implementability.
	
	\subsection{Frequency Attention(FAT)}
	This subsection elaborates on the methodology for deriving Frequency Attention (FAT) from image frequency features and analyzes its effectiveness as an expert kernel weighting allocation mechanism.
	
\begin{figure}[H]
		\centering
		\includegraphics[width=1\textwidth]{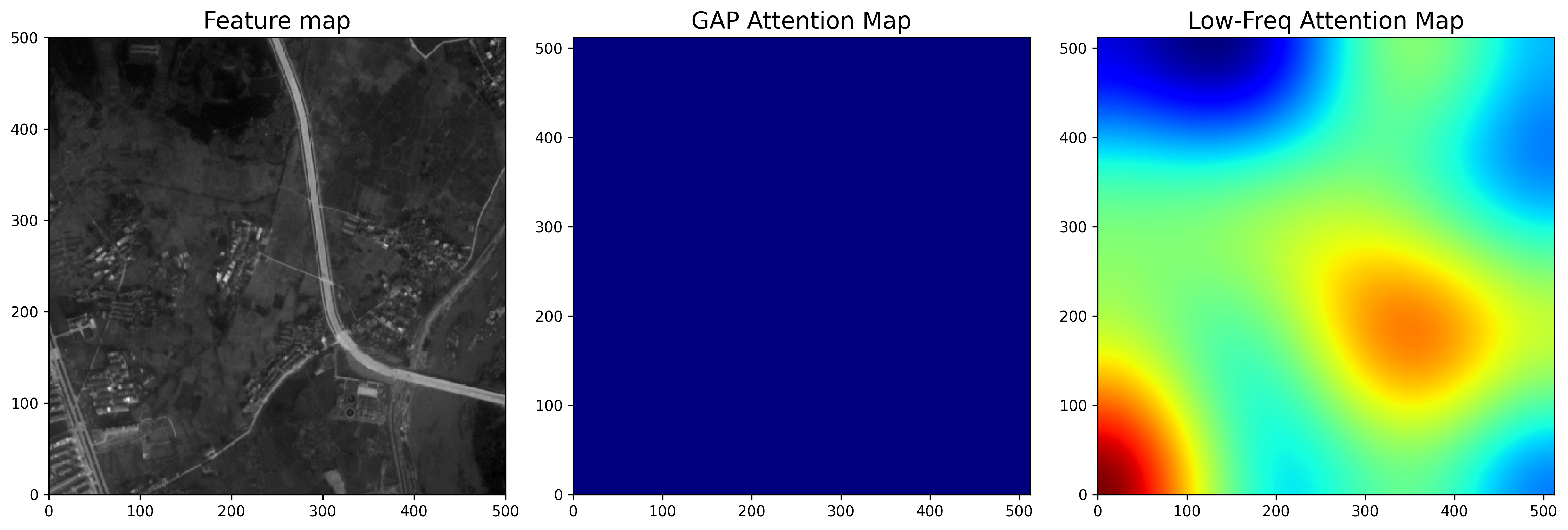}
		\includegraphics[width=1\textwidth]{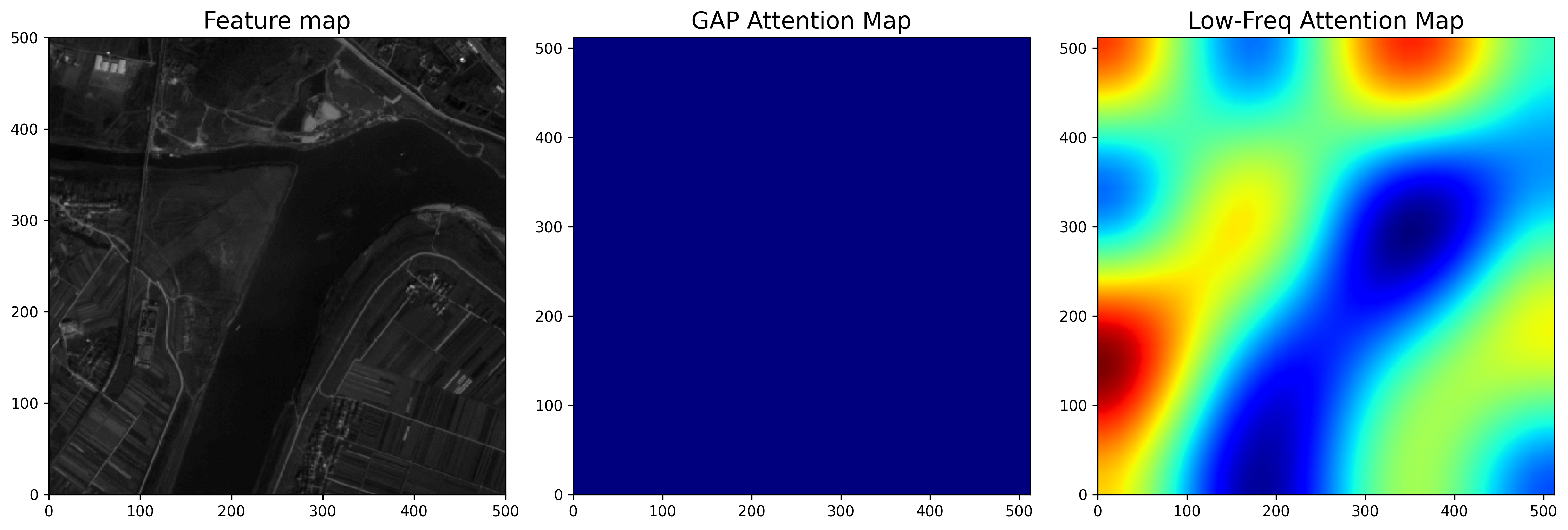}
		\includegraphics[width=1\textwidth]{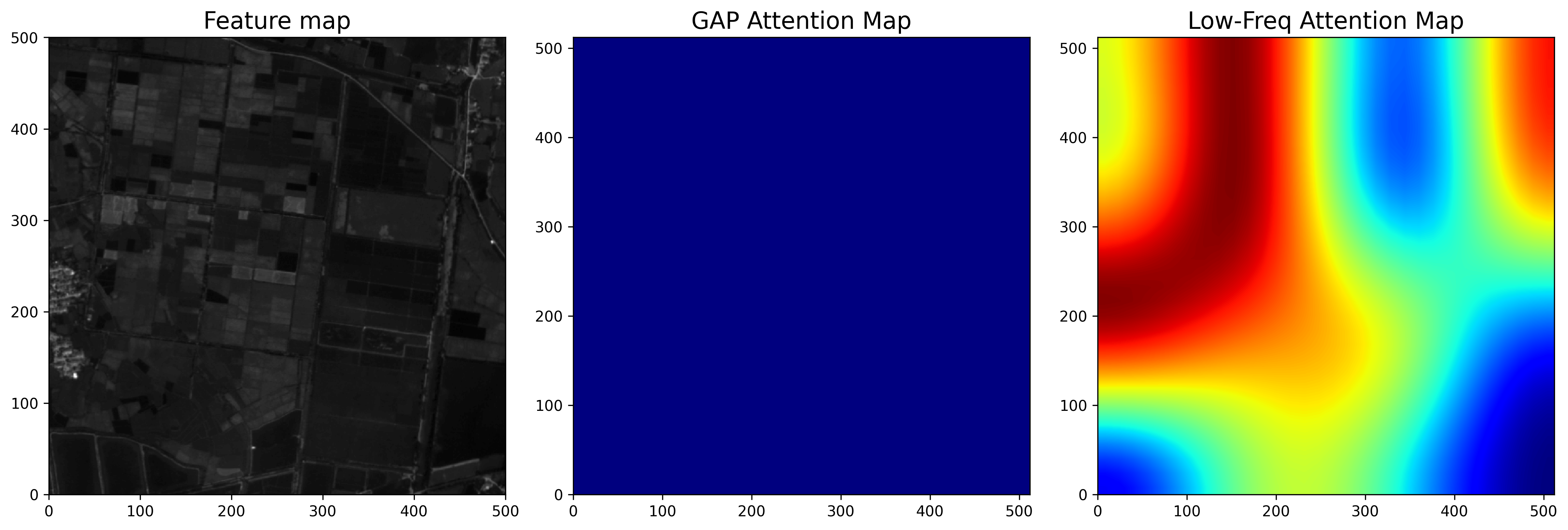}
		\includegraphics[width=1\textwidth]{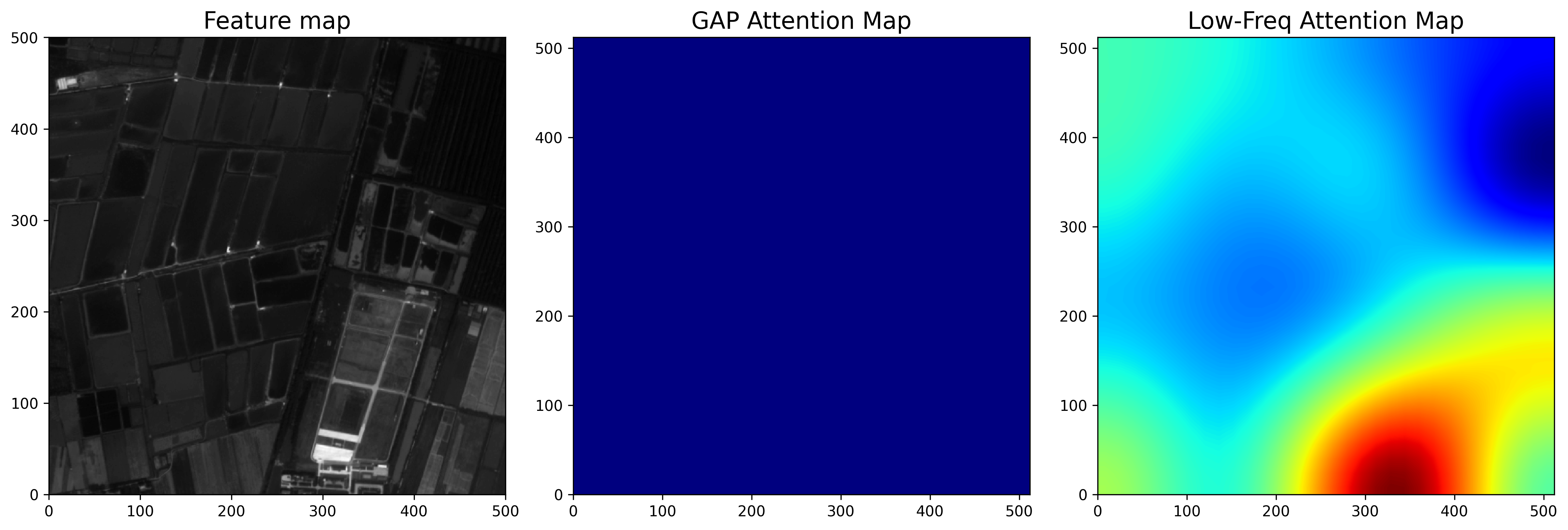}
		\caption{Extract the single-channel image, perform 2D DCT processing, and take the 4x4 size in the upper left corner as the frequency attention. After IDCT, generate the attention heatmap and compare it with the attention heatmap generated by the GAP method.}  
		\label{fig:3-2} 
	\end{figure}

	In the dynamic convolution framework proposed by Yinpeng Chen et al., attention weights for expert kernels are typically obtained through Squeeze-and-Excitation (SE) \citep{Hu2018} operations.These weights are generated by compressing spatial information via GAP, followed by linear layers and activation functions. However, as demonstrated in , GAP-based spatial features mathematically retains only the lowest frequency components while discarding critical information encoded in mid-low and high-frequency bands \citep{Qin2021}, leading to energy dissipation and feature degradation. As illustrated in Figure \ref{fig:3-2}, the left panel displays an input feature map containing diverse geographic objects (e.g., roads, buildings, cropland), visually representing raw geospatial information. The middle panel (GAP attention map) reveals the limitations of global averaging, exhibiting homogeneous attention distribution (blue regions) that fails to focus on semantically critical areas. In sharp contrast, the right panel (low-frequency attention map) shows a phenomenon of local attention concentration (red areas) in different land-cover categories by extracting the medium and low-frequency components and performing IDCT.
	
	The Discrete Cosine Transform (DCT), a linear transformation converting spatial-domain images to the frequency domain, effectively separates low- and high-frequency components. Images of different land-cover categories exhibit Distinct frequency characteristics.Leveraging 2D DCT to capture richer frequency components enables more efficient feature compression. After ignoring the normalization factor that does not affect this work,the 2D DCT formulation is expressed as:
	\begin{equation}
		F(u, v) = \sum_{x=0}^{N-1} \sum_{y=0}^{N-1} f(x, y) \cos \left( \frac{(2x+1)u\pi}{2N} \right) \cos \left( \frac{(2y+1)v\pi}{2N} \right) 
	\end{equation}
	
	where $F \in \mathbb{R}^{H \times W}$ is 2D DCT frequency spectrum.$F(u, v)$ corresponds to the frequency-domain coefficient quantifying energy at frequency $(u, v)$. Applying 2D DCT to an image pixel matrix compresses most of the energy and information into the first element, termed the Direct Current (DC) coefficient, which represents the average brightness of the image block. The remaining coefficients, called Alternating Current (AC) coefficients, represent global characteristics, fine details, and edge variations through multi-frequency representations.
	\\
	\\
	\begin{figure}[H]
		\centering
		\includegraphics[width=1\textwidth]{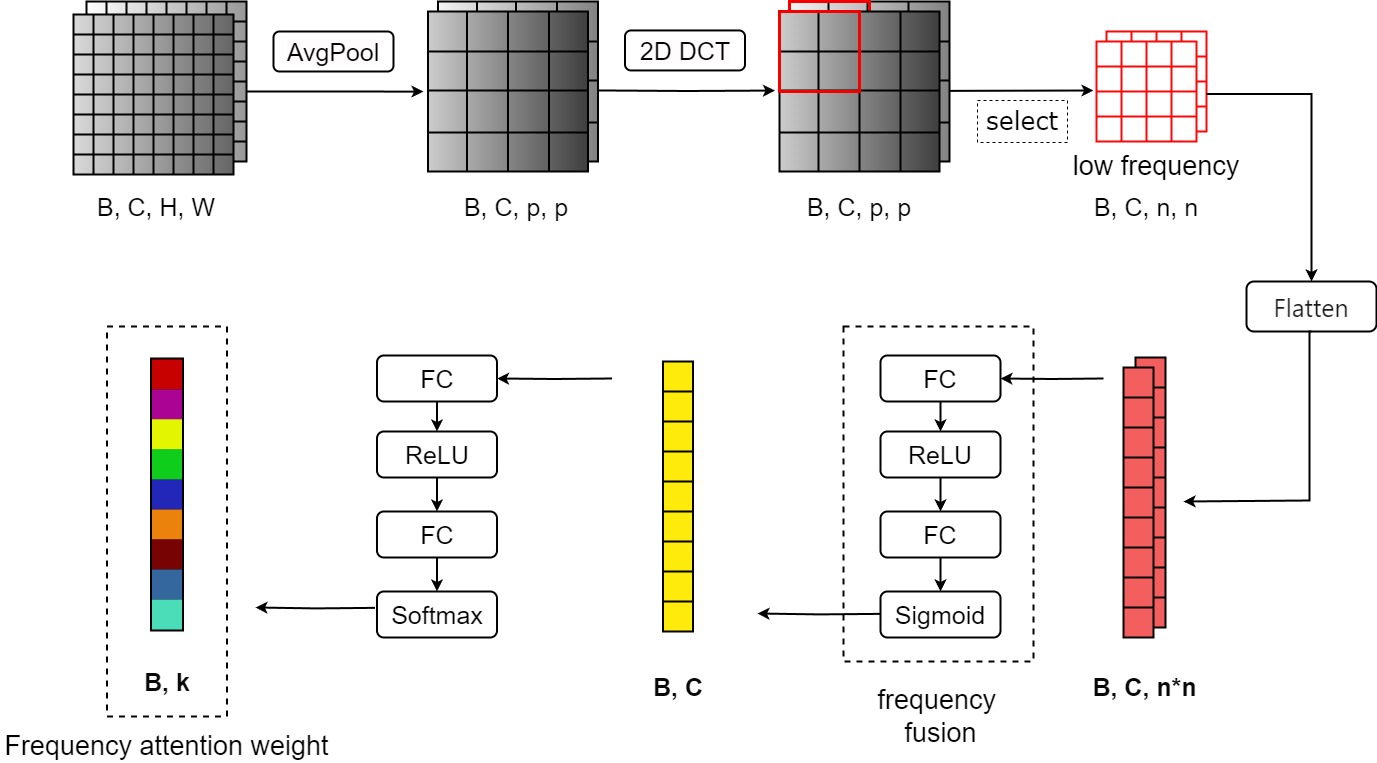}
		\caption{FAT module structure design}  
		\label{fig:3-3} 
	\end{figure}
	\noindent\textbf{FAT Module Design:}
	As illustrated in Figure \ref{fig:3-3}, the FAT Module is designed to extract information-concentrated frequency components from feature maps (with parameter n controlling the extraction scope). This is followed by a frequency feature fusion layer to integrate multi-frequency information within individual channels, and finally a compact network generates dynamic convolution attention weights.
	Specifically, we adaptively pool the feature maps to a predefined size for 2D DCT transformation, where the poolsize (a tunable parameter) is typically set to the minimum dimension of the model's feature maps. The poolsize controls the total computational cost of the 2D DCT transformation, as indicated by the MAdds Formula \ref{MAdds}. After transformation, most energy concentrates in the top-left medium- and low-frequency components. We select the top-left n×n region as the frequency features. These frequency components are flattened into a vector of dimensions [B, C, n×n]. The frequency feature vectors from each channel are fed into the frequency fusion module, processed through a compact network, and aggregated via a sigmoid activation layer to produce 1×1 spatial attention values representing global information of the feature maps, resulting in dimensions [B, C]. Subsequently, a lightweight network (with its size controlled by parameter $r$) processes these values, followed by a softmax layer, to generate frequency attention weights for expert kernel aggregation. The softmax layer ensures that all frequency attention weights sum to 1, normalizing them within the [0, 1] range to stabilize expert kernel weights and improve optimization compatibility.
	\\
	\\
	\textbf{Revisiting 2D DCT from an Energy Response Perspective:}Current studies on DCT primarily emphasize that different frequency components correspond to multi-scale features – for instance, low-frequency components capture global structures while high-frequency components represent edges and noise \citep{Li2025}. This perspective explains the effectiveness of integrating frequency and spatial domains for feature extraction in deep neural networks. However, we argue that frequency-domain characteristics manifest not only in scale-based features but also in energy distribution.As demonstrated in FCANet \citep{Qin2021}, GAP corresponds to a special case of 2D DCT, equivalent to the DC component (zero-frequency term), which reflects the average brightness of an image. The squared magnitude of the DC component further represents the average energy of the input signal. Consider an image patch f(x,y) with a mean pixel value given by:
	\begin{equation}
		\mu = \frac{1}{N^2} \sum_{x=0}^{N-1} \sum_{y=0}^{N-1} I(x, y)
	\end{equation}\\
	The total energy and average energy can be expressed as:
	\begin{equation}
		E_{\text{total}} = \sum_{x=0}^{N-1} \sum_{y=0}^{N-1} f^2(x, y) 
	\end{equation}
	\begin{equation}
		E_{\text{avg}} = \frac{1}{N^2}E_{\text{total}}=\mu^2 \cdot N^2 
	\end{equation}\\
	According to the definition of the DC component of 2D DCT:
	\begin{equation}
		F(0, 0) = \frac{1}{N} \sum_{x=0}^{N-1} \sum_{y=0}^{N-1} f(x,y) = N \cdot \pi 
	\end{equation}\\
	Then the square of the DC component is:
	\begin{equation}
		F^2(0,0)=N^2\cdot\mu^2=E_{\text{avg}}
		\label{formula_1}
	\end{equation}
	
	It can be seen from Formula \ref{formula_1} that the squared magnitude of the DC component equals the average energy of the image. The square of the DC component represents the average level of image energy, while the square of the AC components (the remaining components) indicates the energy causing the undulation of the image. This can be vividly understood as the potential energy of the sea level and the kinetic energy of waves of varying degrees. In SENet\citep{Hu2018}, GAP extracts global information to derive channel attention by focusing on the energy-concentrated DC component. FCANet\citep{Qin2021} employs DCT filters to capture both DC and AC components, fusing multi-frequency features to generate channel attention. This integration of DC and AC energy contributions explains FCANet's superior performance over SENet on ImageNet. Nevertheless, FCANet\citep{Qin2021} applies distinct DCT filters across different channels, leaving individual channels devoid of multi-frequency representations. To address this, we propose performing 2D DCT on each channel to extract multiple frequency components, followed by their fusion to enable intra-channel multi-frequency feature integration. A compact shared network aggregates these frequency features into 1×1 descriptors that encode inter-channel energy disparities and global response intensity, which guide the attention generation process. Unlike GAP-based attention, which coarsely reflects global responses by assuming uniform spatial importance, our FAT module captures frequency-specific energy distributions, allowing selective enhancement or suppression of components. By training a lightweight network to learn the contributions of different frequencies to global features, we achieve finer-grained energy awareness, thereby improving the expressive power of highly compressed global features.
	
	The FAT module, designed to fuse frequency information for attention generation, can replace GAP-based mechanisms in dynamic convolution frameworks. This design ensures compatibility with existing dynamic convolution methods (e.g., ODConv\cite{Li2022}, KernelWarehouse\citep{Li2024}), enhancing both dynamic kernel performance and network representational capacity.

	\section{Experiment}
	In this section, we conduct comparative analyses between existing dynamic convolution methods and our proposed FADConv within CNN backbones, with particular emphasis on cropland segmentation performance to evaluate the effectiveness of our method. Evaluations are performed on two agricultural image datasets to validate FADConv's efficacy. We systematically compare FADConv with multiple dynamic convolution variants at identical backbone positions, assessing performance from multiple perspectives. To ensure fair training conditions and eliminate confounding factors, all experiments are conducted without employing pretrained model fine-tuning, learning rate decay, or other advanced training heuristics, thereby preserving the integrity of comparative results while reserving space for future optimization. All model implementations and dynamic convolution modules are reproduced strictly based on published methodologies and open-source codebases to guarantee experimental fairness
	\subsection{Experimental Setup and Evaluation Metrics}
	The experiments were conducted on a single NVIDIA® GeForce® RTX 4090 GPU with 24GB memory. Given the focus of this study on cultivated land extraction for addressing Non-agriculturalization issues, the evaluation prioritizes precision, F1-score, recall, and IoU metrics for the cultivated land category, while secondary attention is given to other land cover categories.
	For the binary classification on the Hi-CNA dataset\citep{Tong2020}, the metrics include accuracy, precision, recall, F1-score, and IoU. For the six-class classification on the GID dataset\citep{sun2024}, the metrics comprise overall accuracy (OA), mean IoU (mIoU), and cultivated land-specific metrics (accuracy, precision, recall, F1-score, and IoU). The mathematical formulations are defined as follows:
	\begin{equation}
		{Accuracy} = \frac{TP + TN}{TP + TN + FP + FN}
	\end{equation}
	
	\begin{equation}
		{Precision} = \frac{TP}{TP + FP}
	\end{equation}
	
	\begin{equation}
		{Recall} = \frac{TP}{TP + FN}
	\end{equation}
	
	\begin{equation}
		F1 = \frac{2}{Precision^{-1}+Recall^{-1}}
	\end{equation}
	
	\begin{equation}
		{IoU} = \frac{TP}{TP + FP + FN}
	\end{equation}
	
	\begin{equation}
		{OA} = \frac{\sum_{i=1}^{k} (TP_i + TN_i)}{\sum_{i=1}^{k} (TP_i + TN_i + FP_i + FN_i)}
	\end{equation}
	
	\begin{equation}
		{mIoU} = \frac{1}{k} \sum_{i=1}^{k} \text{IoU}_i
	\end{equation}
	
	\subsection{Descriptions of Datasets}
	This paper focuses on the issue of non-agriculturalization and mainly addresses the semantic segmentation of cultivated land areas. Two datasets are used for experimental evaluation: the GID dataset\citep{Tong2020} is the primary experimental dataset, and the Hi-CNA dataset\citep{sun2024} is used for auxiliary experiments. The basic information of the two datasets is as follows:
	
	(1) The GID dataset\citep{Tong2020} is based on GF-2 high-resolution remote sensing images with a spatial resolution of 0.8m. The GID dataset contains 150 GF-2 satellite remote sensing images with pixel-level annotations, each with a size of 6800x7200 pixels. The classification categories include five land cover types: buildings, cropland, forests, grasslands, and water bodies. The dataset covers over 60 different cities in China, and each image is clear and of high quality without cloud or fog obstruction. The overall image coverage exceeds 50,000 square kilometers of geographical area. In the experiments of this paper, the images are non-overlappingly segmented into 512x512 pixel sizes, resulting in a total of 27,300 images. Twenty percent of them are randomly selected as the test set, and eighty percent as the training set.
	
	(2)The Hi-CNA dataset\citep{sun2024} is derived from multispectral GF-2 fused images with a spatial resolution of 0.8m, including four bands: blue, green, red, and near-infrared. The study area covers provinces such as Hebei, Shanxi, Shandong, and Hubei in China, with a total area of over 1,100 square kilometers. The dataset annotates two types of areas: cultivated land and non-cultivated land, covering two time periods: the first period from 2015 to 2017, and the second period from 2020 to 2022. All images are cropped into 512x512 pixel slices. The Hi-CNA dataset contains a total of 13,594 images, with sixty percent as the training set, twenty percent as the validation set, and twenty percent as the test set.
	
	\subsection{Experiment of GID dataset}
	Our experiments were primarily conducted on the GID dataset\citep{sun2024}, which contains 27,300 images for training and testing. The dataset includes 6 classes (with background as one category), and we focus on evaluating metrics for the cropland class. 
	\\
	\\
	\textbf{CNN Backbones:}We choose MobileNetV2\citep{Howard2018}, DeeplabV3+\citep{chen2018_2}, and ResNet\citep{He2016} families for the experiments. Specifically, it includes MobileNetV2 as the encoder and U-net\citep{Ronneberger2015} structure as the decoder; DeeplabV3+ with Resnet18 as the backbone, retaining the original decoder structure; and both ResNet18 and ResNet34 as encoders with Unet as the decoder.
	\\
	\\
	\textbf{Experimental Setup:}In the experiment, we compare FADConv with related methods to demonstrate its effectiveness. The two dynamic convolution methods most relevant to our work are DYConv\citep{chen_y2020} and ODConv\citep{Li2022}. DYConv is the pioneering work in dynamic convolution, while ODConv extends the dimension of dynamic attention and has achieved good results on the ImageNet dataset. Therefore, in the MobileNet and Deeplab backbone networks, we choose them as the main reference methods. Additionally, we incorporate the KernelWarehouse (KW)\citep{Li2024} method in the ResNets backbone for reference. KW innovates the dynamic combination of convolution kernels by changing the previous parallel multi-kernel approach and using a kernel warehouse sharing mechanism. We conduct experiments with these dynamic convolution methods under the same conditions. In the MobileNet backbone network encoder, we replace both depthwise convolution and pointwise convolution with each dynamic convolution method. For the encoder part of Deeplab and ResNets, except for the first layer structure, we replace them with each dynamic convolution method for comparative experiments. The experiments are conducted on the GID dataset with a unified training configuration: all experiments set the learning rate to 0.0001, weight decay to 1e-4, batch size to 24, and epoch to 50. No sample augmentation was used.The loss function adopts CrossEntropy loss. We use b\texttimes to represent the number of parallel convolution kernels or the convolution parameter budget relative to normal convolution for each dynamic convolution method.
	\begin{table}[H]
		\centering
		\caption{The scores obtained by testing various dynamic convolution methods on MobilenetV2 and DeeplabV3+ using the GID dataset. Best results are bolded.}
		\label{tab:model_comparison}
		\footnotesize
		\begin{tabular}{lllllllll}
			\toprule
			\multicolumn{1}{l}{\textbf{Model}} & 
			\multicolumn{1}{c}{\textbf{OA}} & 
			\multicolumn{1}{c}{\textbf{Acc}} & 
			\multicolumn{1}{c}{\textbf{P}} & 
			\multicolumn{1}{c}{\textbf{R}} & 
			\multicolumn{1}{c}{\textbf{F1}} & 
			\multicolumn{1}{c}{\textbf{IoU}} & 
			\multicolumn{1}{c}{\textbf{Params}} & 
			\multicolumn{1}{l}{\textbf{MAdds}} \\
			\midrule
			\multicolumn{9}{l}{\textbf{MobilenetV2}} \\
			Baseline & 78.3 & 88.1 & 80.0 & 81.6 & 80.8 & 67.8 & 3.67M & 3.23G \\
			+ DYConv(4\texttimes) & 78.8 & 88.5 & 80.6 & 82.1 & 81.3 & 68.6 & 10.95M & +9.44M \\
			+ ODConv(4\texttimes) & 78.9 & 88.3 & 79.1 & \uline{83.9} & 81.5 & 68.7 & 11.97M & +40.71M \\
			+ FDConv(4\texttimes, p=16) & \uline{80.1} & \uline{89.2} &\uline {80.9} & \textbf{84.7} & \uline{82.7} & \uline{70.5} & 10.95M & +138.70M \\
			+ FDConv(4\texttimes, p=32) & \textbf{80.4} & \textbf{89.4} & \textbf{82.4} & 83.4 & \textbf{82.9} & \textbf{70.7} & 10.95M & +1.04G \\
			\midrule
			\multicolumn{9}{l}{\textbf{DeeplabV3+}} \\
			Baseline & 81.9 & 90.4 & 83.5 & 85.4 & 84.5 & 73.1 & 16.53M & 31.56G \\
			+ DYConv(4\texttimes) & 82.7 & 91.0 & 84.8 & 86.0 & 85.4 & 74.5 & 62.33M & +60.95M \\
			+ ODConv(4\texttimes) & 82.8 & 90.9 & 83.0 & \textbf{88.5} & 85.6 & 74.9 & 62.84M & +229.10M \\
			+ FDConv(4\texttimes, p=16) & \textbf{83.0} & \textbf{91.1} & \textbf{83.6} & 88.1 & \textbf{85.8} & \textbf{75.1} & 62.33M & +119.71M \\
			+ FDConv(4\texttimes, p=32) & \textbf{83.0} & \textbf{91.1} & \uline{83.5} & \uline{88.3} & \textbf{85.8} & \textbf{75.1} & 62.33M & +530.92M \\
			\bottomrule
		\end{tabular}
		\vspace{0.3cm}
		
		\footnotesize \textit{Note:} OA denotes the overall pixel-level accuracy. Acc, P, R, F1, and IoU represent the pixel-level accuracy, precision, recall, F1 score, and intersection over union for the arable land class, respectively.Params = Number of parameters, MAdds = Multiply-Add operations. 
		The best results are shown in \textbf{bold}, while the second-best results are \uline{underlined}. In the MAdds column, "+" indicates the increase in computational cost relative to the corresponding Baseline.
		\label{table1} 
	\end{table}
	
\begin{figure}[H]
		\centering
		\subcaptionbox{}{\includegraphics[width=0.15\textwidth]{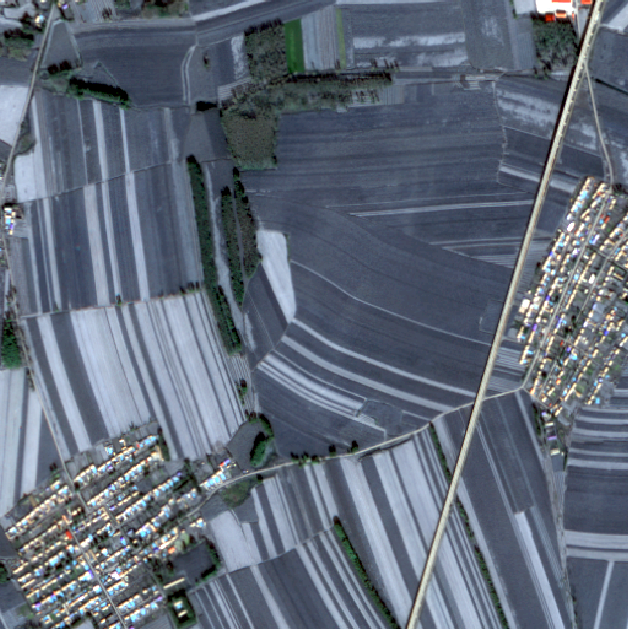}}\hfill
		\subcaptionbox{}{\includegraphics[width=0.15\textwidth]{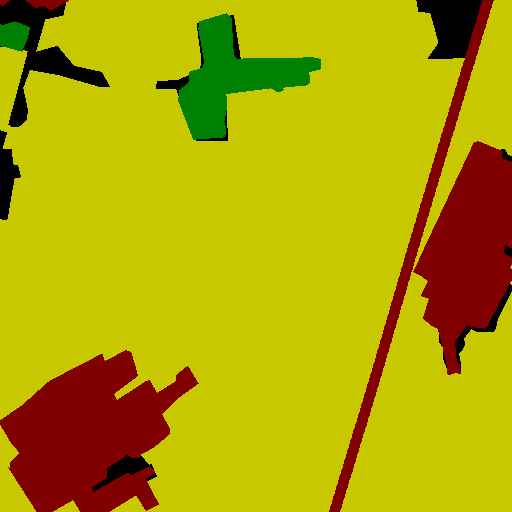}}\hfill
		\subcaptionbox{}{\includegraphics[width=0.15\textwidth]{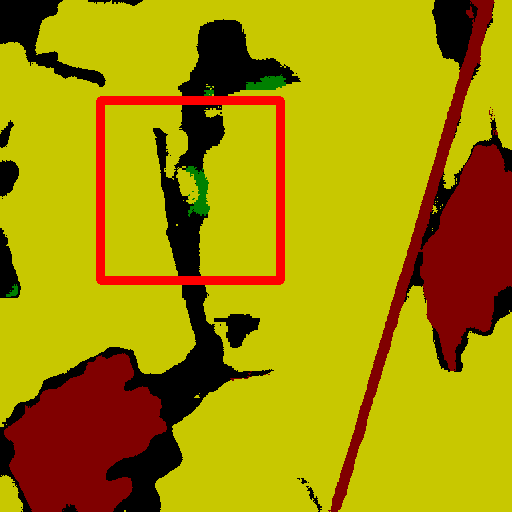}}\hfill
		\subcaptionbox{}{\includegraphics[width=0.15\textwidth]{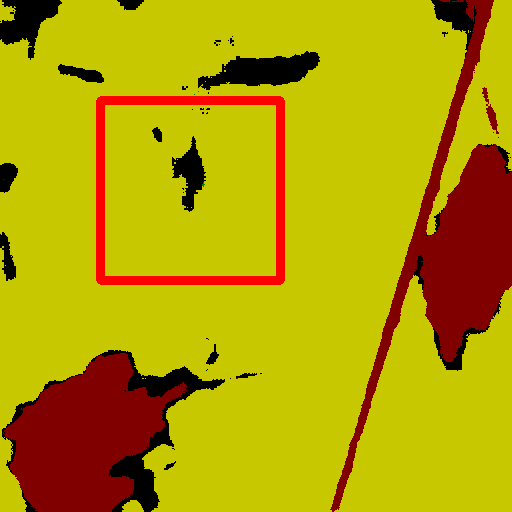}}\hfill
		\subcaptionbox{}{\includegraphics[width=0.15\textwidth]{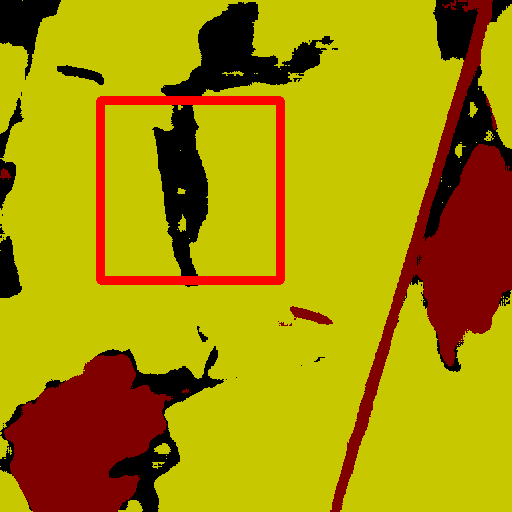}}\hfill
		\subcaptionbox{}{\includegraphics[width=0.15\textwidth]{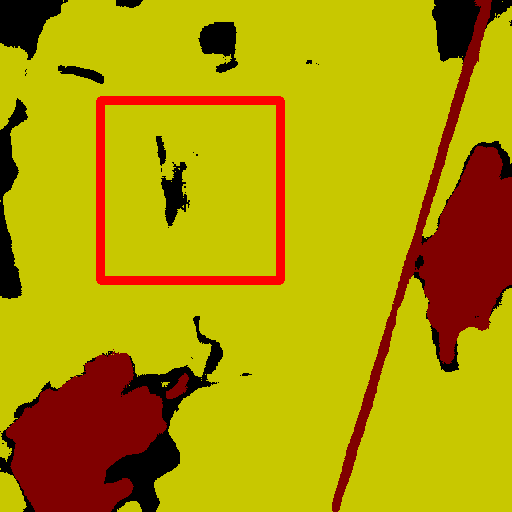}}
		\setcounter{subfigure}{0} 
		\subcaptionbox{}{\includegraphics[width=0.15\textwidth]{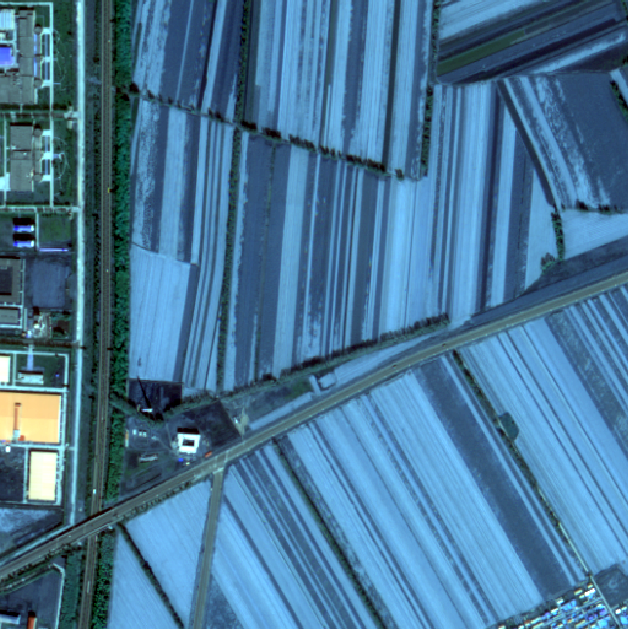}}\hfill
		\subcaptionbox{}{\includegraphics[width=0.15\textwidth]{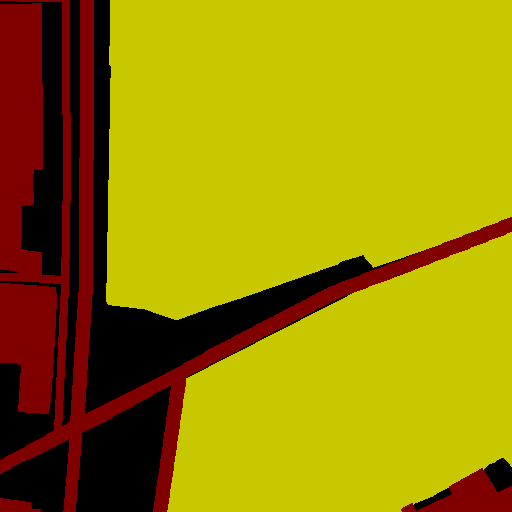}}\hfill
		\subcaptionbox{}{\includegraphics[width=0.15\textwidth]{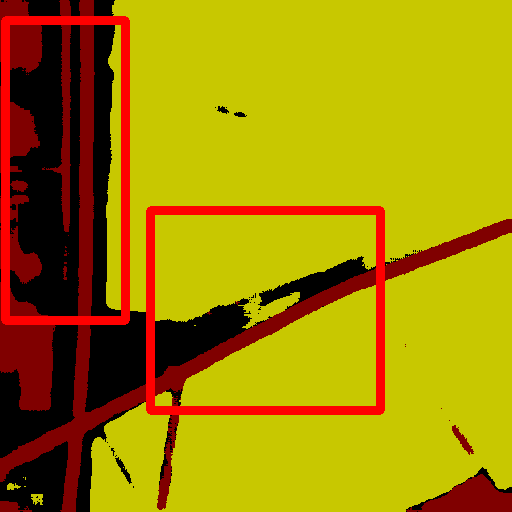}}\hfill
		\subcaptionbox{}{\includegraphics[width=0.15\textwidth]{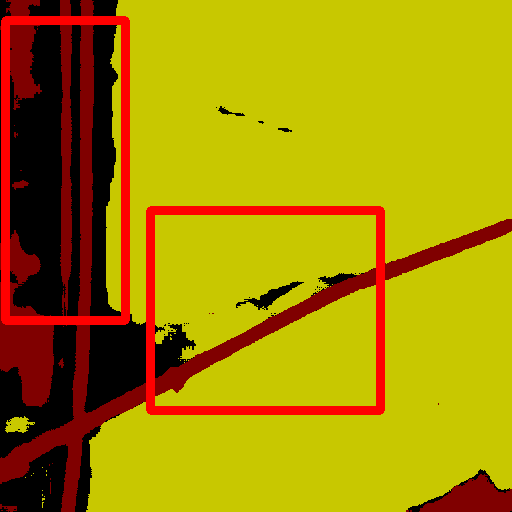}}\hfill
		\subcaptionbox{}{\includegraphics[width=0.15\textwidth]{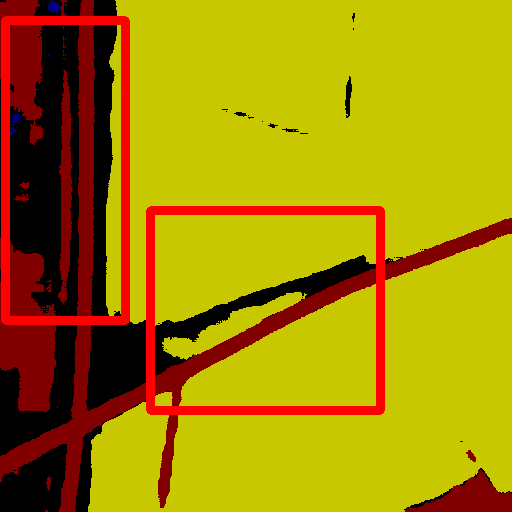}}\hfill
		\subcaptionbox{}{\includegraphics[width=0.15\textwidth]{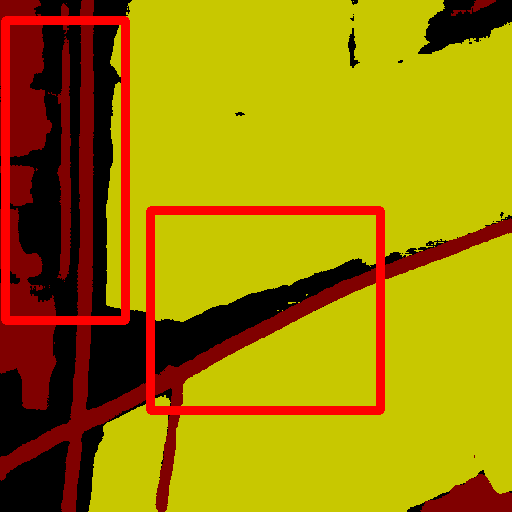}}
		\setcounter{subfigure}{0} 
		\subcaptionbox{}{\includegraphics[width=0.15\textwidth]{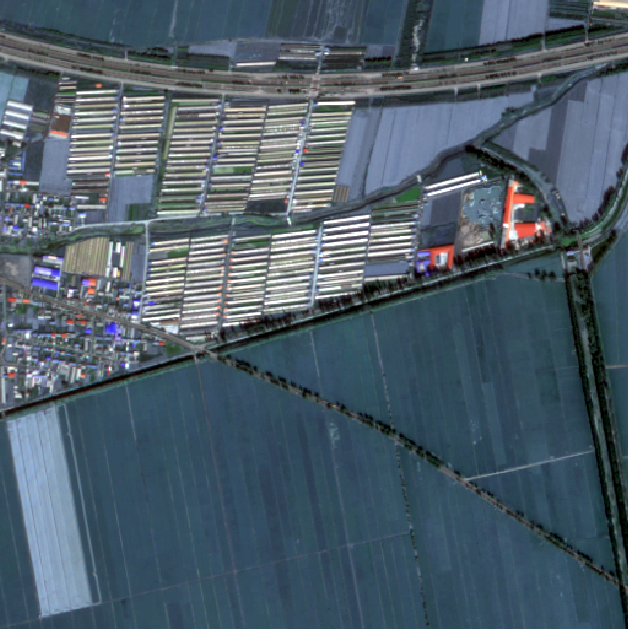}}\hfill
		\subcaptionbox{}{\includegraphics[width=0.15\textwidth]{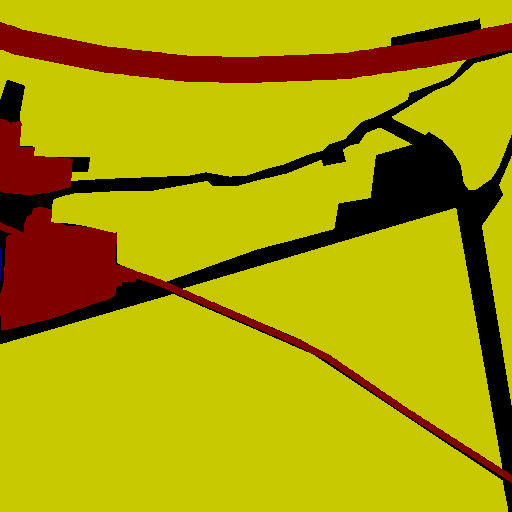}}\hfill
		\subcaptionbox{}{\includegraphics[width=0.15\textwidth]{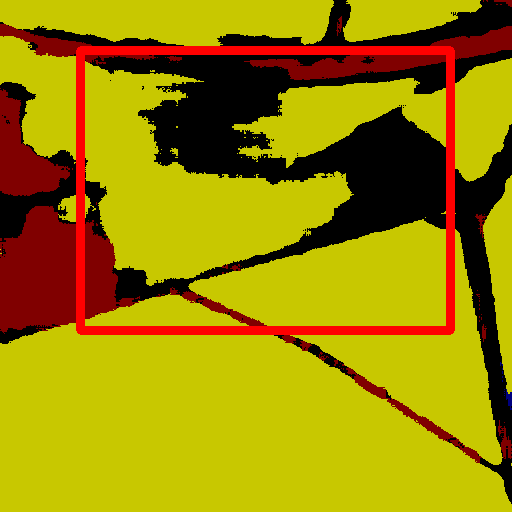}}\hfill
		\subcaptionbox{}{\includegraphics[width=0.15\textwidth]{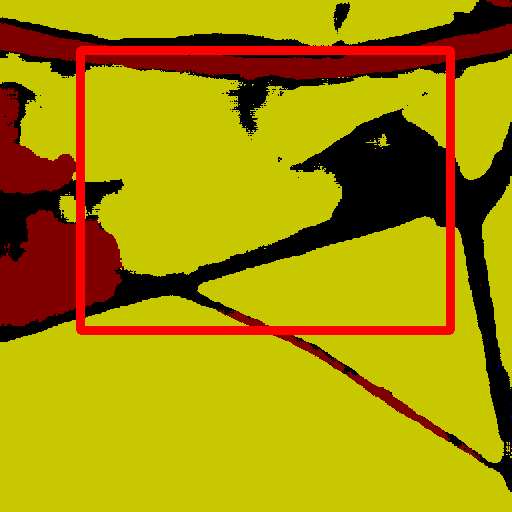}}\hfill
		\subcaptionbox{}{\includegraphics[width=0.15\textwidth]{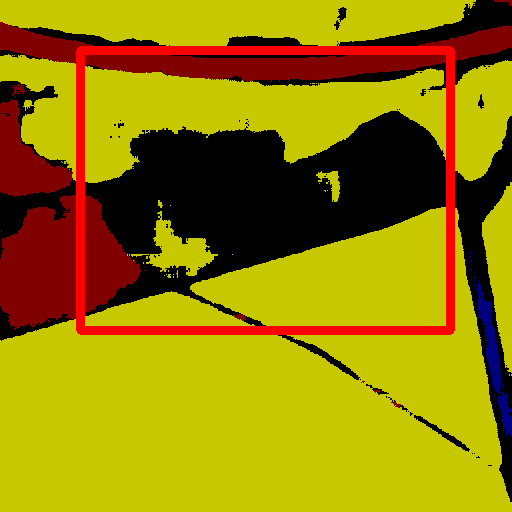}}\hfill
		\subcaptionbox{}{\includegraphics[width=0.15\textwidth]{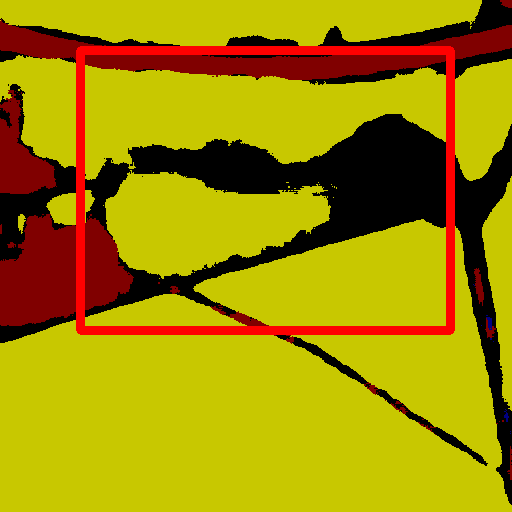}}
		\setcounter{subfigure}{0} 
		\subcaptionbox{}{\includegraphics[width=0.15\textwidth]{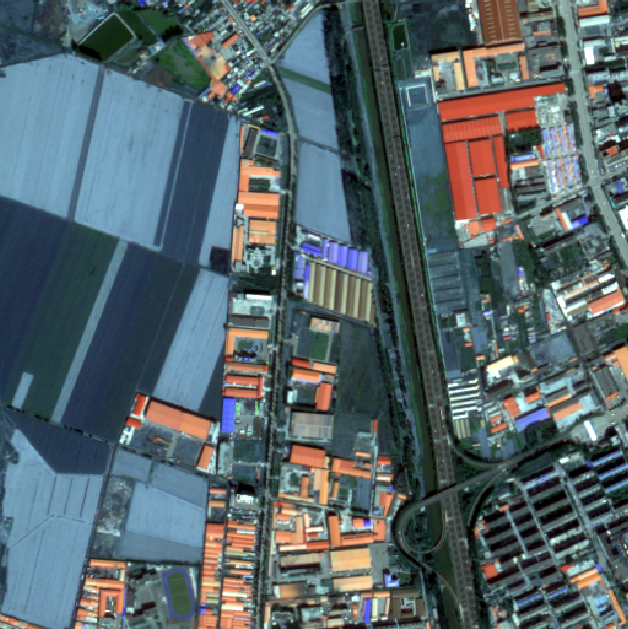}}\hfill
		\subcaptionbox{}{\includegraphics[width=0.15\textwidth]{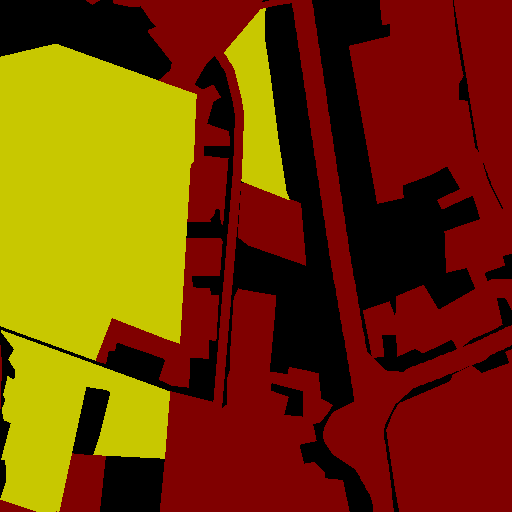}}\hfill
		\subcaptionbox{}{\includegraphics[width=0.15\textwidth]{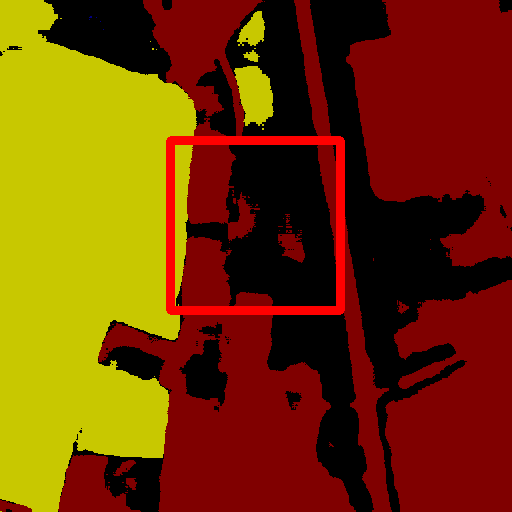}}\hfill
		\subcaptionbox{}{\includegraphics[width=0.15\textwidth]{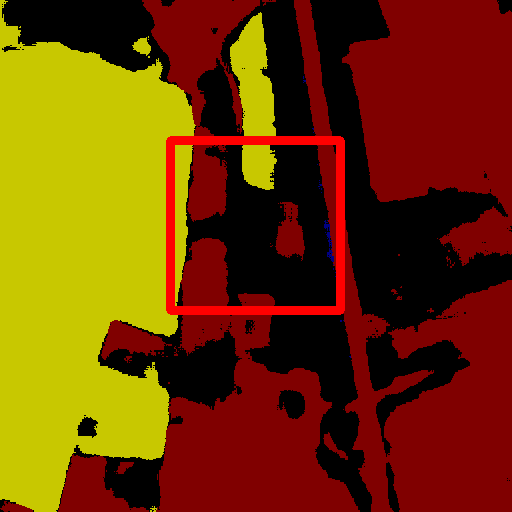}}\hfill
		\subcaptionbox{}{\includegraphics[width=0.15\textwidth]{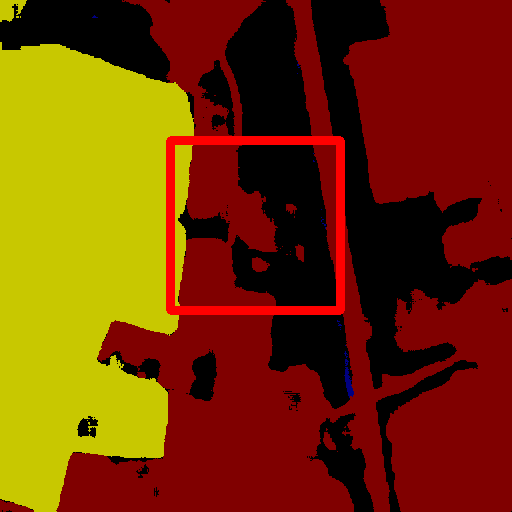}}\hfill
		\subcaptionbox{}{\includegraphics[width=0.15\textwidth]{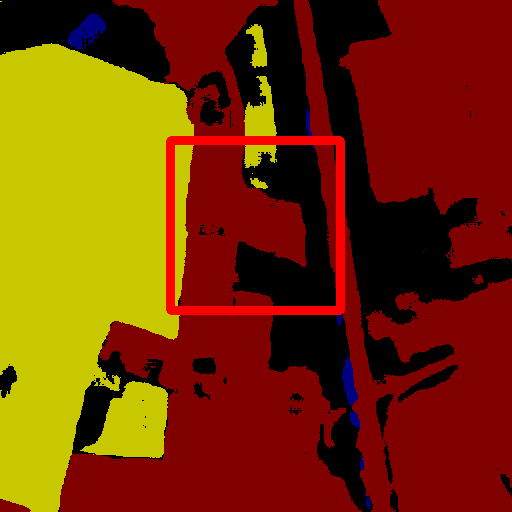}}
		\setcounter{subfigure}{0} 
		\subcaptionbox{}{\includegraphics[width=0.15\textwidth]{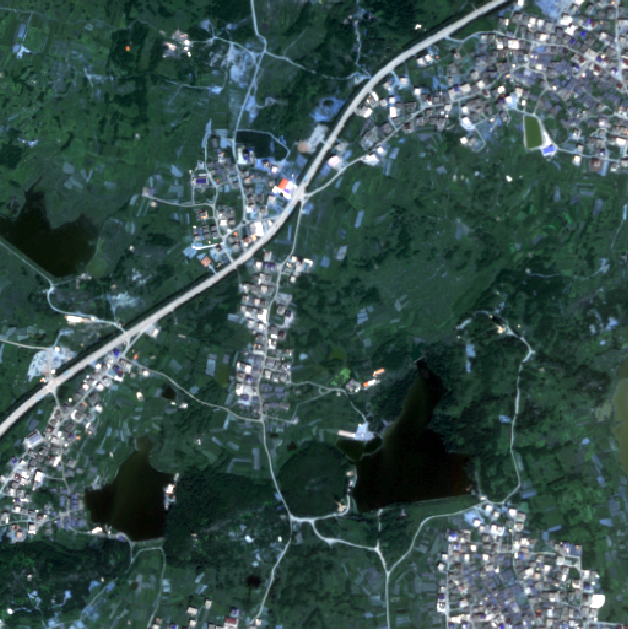}}\hfill
		\subcaptionbox{}{\includegraphics[width=0.15\textwidth]{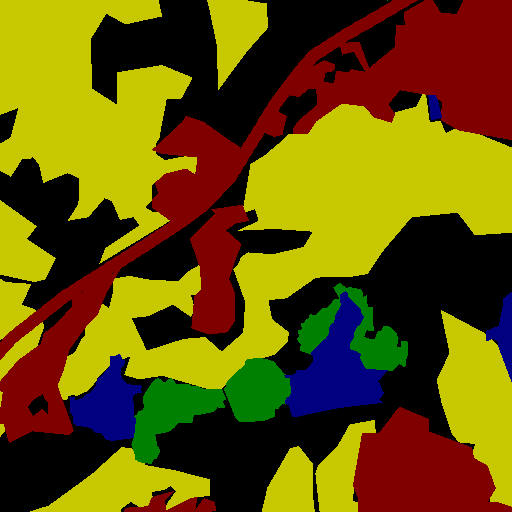}}\hfill
		\subcaptionbox{}{\includegraphics[width=0.15\textwidth]{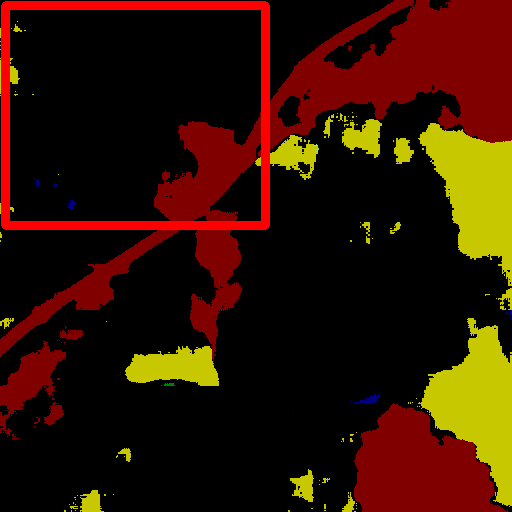}}\hfill
		\subcaptionbox{}{\includegraphics[width=0.15\textwidth]{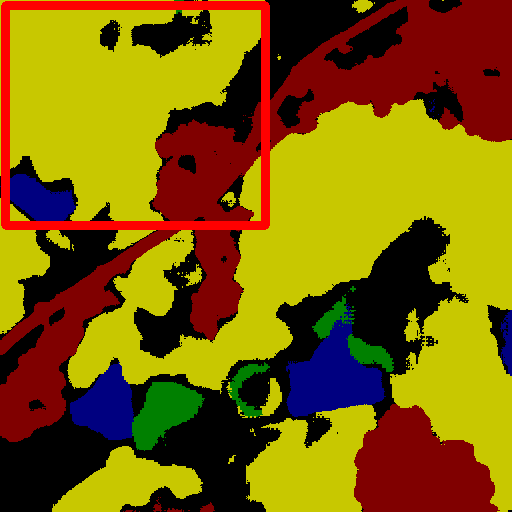}}\hfill
		\subcaptionbox{}{\includegraphics[width=0.15\textwidth]{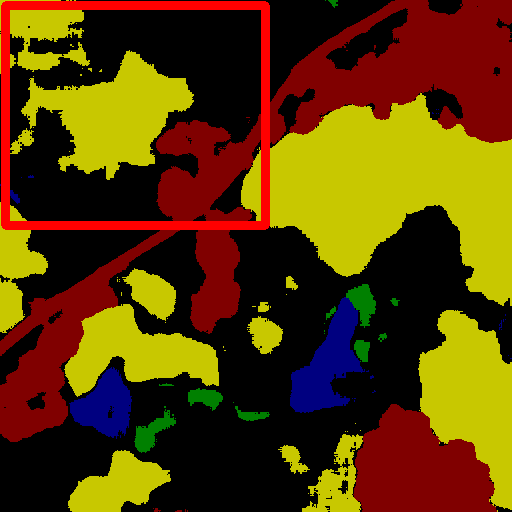}}\hfill
		\subcaptionbox{}{\includegraphics[width=0.15\textwidth]{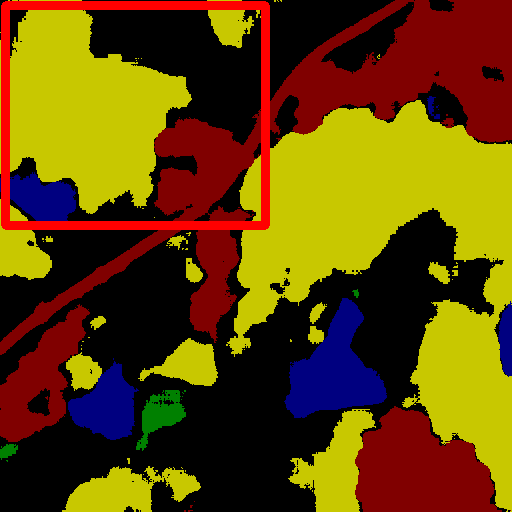}}
		
		\caption{The segmentation results on the GID dataset after applying the dynamic convolution method to ResNet34.
			(a) Row image (b) Label ,(c) +DYConv ,(d) +ODConv ,(e) +KW ,(f) +FADConv(ours)}
		\label{fig:4-1}
	\end{figure}

	\noindent\textbf{Results Comparison on MobileNet and DeepLab:}
	We conducted experiments using MobileNet and DeepLab as backbones to validate FADConv's performance in lightweight architectures and multi-scale feature fusion. Test results (Table \ref{table1}) demonstrate that FADConv, which adjusts dynamic kernel composition via frequency-based attention, effectively balances parameter efficiency and representational capacity while extracting global features in both lightweight and multi-scale structures. In MobileNetV2 – which contains numerous depthwise and pointwise convolutional layers – FADConv (4\texttimes) incurs higher computational costs due to 2D DCT operations. By reducing the pool size to 16, it achieves significant improvements (1.9\% F1-score and 2.7\% IoU over baseline) with only a 4.3\% increase in computational overhead, outperforming other dynamic convolution methods. For DeepLabV3+, FADConv (4x) also delivers the best performance: with pool size=16, it improves F1-score/IoU by 1.3\%/2.0\% compared to the baseline network while adding 119.71M MAdds. These results validate the effectiveness of frequency-aware attention even when dynamic convolutions are applied to depthwise separable convolutions and atrous convolutions.
	Note: We did not add the comparison experimental data of KW in the data tables of MobileNet and DeepLab because under the fixed experimental learning rate condition, the MobileNet model with the KW method added showed a difficult-to-converge situation. fourteen
	However, due to the principle that KW shares the same convolutional kernels, its effect on DeepLab is equivalent to that on ResNet. Therefore, no experiments were conducted using KW in DeepLab.
	\begin{table}[H]
		\centering
		\caption{The scores obtained by testing various dynamic convolution methods on ResNet18 and ResNet34 using the GID dataset. Best results are bolded.}
		\label{tab:crack_results}
		\footnotesize
		\begin{tabular}{lllllllll} 
			\toprule
			\multicolumn{1}{l}{\textbf{Model}} & 
			\multicolumn{1}{c}{\textbf{OA}} & 
			\multicolumn{1}{c}{\textbf{Acc}} & 
			\multicolumn{1}{c}{\textbf{P}} & 
			\multicolumn{1}{c}{\textbf{R}} & 
			\multicolumn{1}{c}{\textbf{F1}} & 
			\multicolumn{1}{c}{\textbf{IoU}} & 
			\multicolumn{1}{c}{\textbf{Params}} & 
			\multicolumn{1}{l}{\textbf{MAdds}} \\
			\midrule
			\multicolumn{9}{l}{\textbf{ResNet18}} \\
			Baseline & 80.9 & 89.5 & 81.3 & 85.5 & 83.3 & 71.5 & 17.97M & 50.68G \\
			+DYConv(4\texttimes) & 81.6 & 90.1 & 83.0 & 85.0 & 84.0 & 72.5 & 44.06M & +34.75M \\
			+ODConv(4\texttimes) & 81.1 & 89.9 & 83.4 & 83.7 & 83.7 & 71.7 & 44.27M & +96.66M \\
			+KW(4x) & 81.4 & 89.9 & \uline{83.7} & 83.3 & 83.5 & 71.7 & 45.21M & +0.56G \\
			+FDConv(4\texttimes, p=16) & \textbf{82.5} & \uline{90.6} & 82.3 & \textbf{88.3} & \uline{85.2} & \uline{74.2} & 44.07M & +58.87M \\
			+FDConv(4\texttimes, p=32) & \textbf{82.5} & \textbf{90.9} & \textbf{84.4} & 
			\uline{86.1} & \textbf{85.3} & \textbf{74.3} & 44.07M & +227.6M \\
			\midrule
			\multicolumn{9}{l}{\textbf{ResNet34}} \\
			Baseline & 81.9 & 90.4 & 84.0 & 85.0 & 84.5 & 73.1 & 28.08M & 60.35G \\
			+DYConv(4\texttimes) & 82.3 & 90.7 & \textbf{84.2} & 85.7 & 84.9 & 73.9 & 84.60M & +75.27M \\
			+ODConv(4\texttimes) & 82.1 & 90.5 & 83.4 & 86.1 & 84.7 & 73.5 & 85.05M & +204.6M \\
			+KW(4x) & 82.6 & 90.8 & \textbf{84.2} & 86.2 & 85.2 & 74.2 & 87.31M & +1.29G \\
			+FDConv(4\texttimes, p=16) & \textbf{83.0} & \textbf{91.2} & 82.7 & \textbf{89.9} & \textbf{86.2} & \textbf{75.7} & 84.62M & +129.81M \\
			+FDConv(4\texttimes, p=32) & \uline{82.9} & \textbf{91.2} & 84.1 & \uline{87.6} & \uline{85.8} & \uline{75.1} & 84.62M & +511.40M \\
			\bottomrule
		\end{tabular}
		\vspace{0.3cm}
		\label{table2}
	\end{table}
	\noindent\textbf{Results Comparison on ResNets:}
	We conducted extensive experiments on ResNets to investigate FADConv’s efficacy in CNNs. First, we validated FADConv’s effectiveness across ResNet18 and ResNet34 to assess its scalability with network depth. In addition to DY-Conv and ODConv, we included KernelWarehouse (KW) as a reference method. While FADConv shares structural similarities with DY-Conv and ODConv in kernel design, KW introduces a distinct attention-based kernel-sharing mechanism. As shown in Table \ref{table2}, FADConv (4\texttimes) achieves best performance on the GID dataset with lower computational and parametric costs. For ResNet18, FADConv (4\texttimes) with poolsize=32 delivers the best results (2.0\% F1-score and 2.8\% IoU gains over baseline), A similar effect was observed when poolsize was set to 16. The best results were achieved in ResNets34 when the poolsize was set to 16 (1.7\% F1-score and 2.6\% IoU gains). These results demonstrate FADConv’s superiority over existing dynamic convolution methods without excessive parameter or computation overhead. Segmentation results for ResNet34 with dynamic convolutions are visualized in Figure \ref{fig:4-1}, where FADConv exhibits higher accuracy. Notably, inconsistent poolsize performance trends suggest dataset-specific optimization, which we further explore in the Hi-CNA experiments.

\begin{figure}[H]
		\centering
		\includegraphics[width=0.9\textwidth]{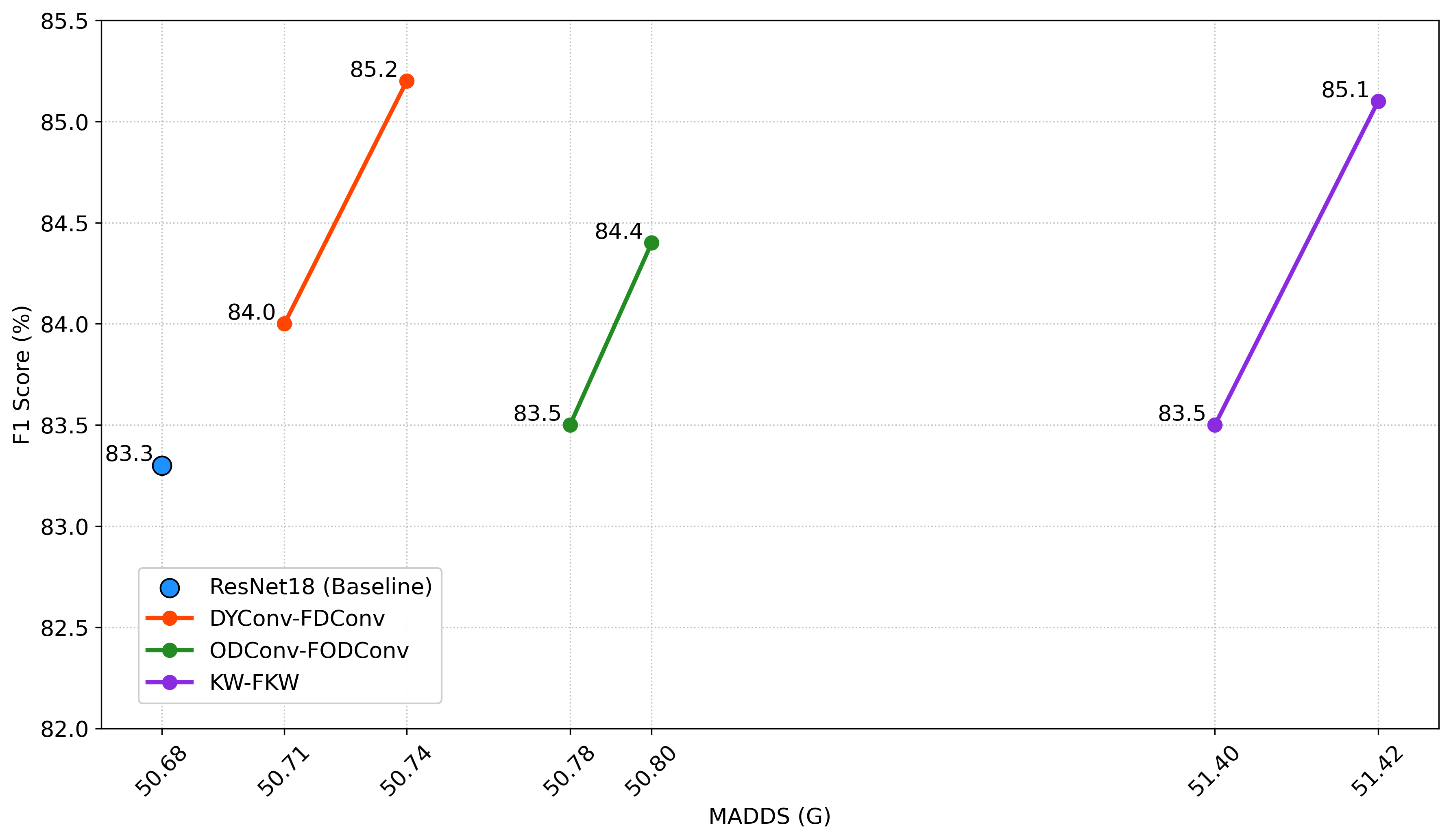}
		\caption{On ResNet18, the GAP method in various dynamic convolution methods was replaced with FAT and tested on the GID dataset. The vertical axis represents the F1 Score, and the horizontal axis represents MAdds, with the unit being G.}  
		\label{fig:4-2} 
	\end{figure}

	It is worth noting that while ODConv and KW perform well on large-scale datasets like ImageNet/COCO, they show weaker generalization on smaller or imbalanced datasets. For example, on ResNet18, their F1-scores lag behind FADConv by 1.8\% and 1.8\%, respectively. This indicates limitations in their attention mechanisms, whereas FADConv’s frequency-aware design enhances generalization. 
	To validate this, we replaced GAP-based attention in DYConv, ODConv, and KW with our FAT module. Results (Table \ref{table3} and Figure \ref{fig:4-2}) confirm FAT’s compatibility and effectiveness, improving generalization without significant computational penalties.
	Training and loss curves for ResNet34 (baseline vs. FADConv) reveal comparable training performance but superior validation results with FADConv.This suggests that deeper networks sufficiently learn feature patterns, yet frequency-aware dynamic convolution better handles data diversity.Unlike methods relying on parameter expansion, FADConv leverages frequency features to process complex patterns adaptively, aligning with the core advantage of dynamic convolution – balancing model capacity and generalization.
	\begin{table}[htbp]
		\centering
		\caption{The traditional dynamic convolution method for obtaining attention is replaced by using the FAT module. The value of p is set to 16. Best results are bolded. }
		\footnotesize
		\begin{tabular}{lllllllll}
			\toprule
			\multicolumn{1}{l}{\textbf{Model}} & 
			\multicolumn{1}{c}{\textbf{OA}} & 
			\multicolumn{1}{c}{\textbf{Acc}} & 
			\multicolumn{1}{c}{\textbf{P}} & 
			\multicolumn{1}{c}{\textbf{R}} & 
			\multicolumn{1}{c}{\textbf{F1}} & 
			\multicolumn{1}{c}{\textbf{IoU}} & 
			\multicolumn{1}{c}{\textbf{Params}} & 
			\multicolumn{1}{l}{\textbf{MAdds}} \\
			\midrule
			\multicolumn{9}{l}{\textbf{ResNet18}} \\
			Baseline & 80.9 & 89.5 & 81.3 & 85.5 & 83.3 & 71.5 & 17.97M & 50.68G \\
			\midrule
			+DYConv(4\texttimes) & 81.6 & 90.1 & \textbf{83.0} & 85.0 & 84.0 & 72.5 & 44.06M & +34.75M \\
			+FDConv(4\texttimes) & \textbf{82.5} & \textbf{90.6} & 82.3 & \textbf{88.3} & \textbf{85.2} & \textbf{74.2} & 44.06M & +58.87M \\
			\midrule
			+ODConv(4\texttimes) & 81.1 & 89.9 & 83.4 & 83.7 & 83.6 & 71.7 & 44.27M & +96.66M \\
			+FODConv(4\texttimes) & \textbf{81.8} & \textbf{90.4} & \textbf{83.5} & \textbf{85.2} & \textbf{84.4} & \textbf{73.0} & 44.27M & +120.78M \\
			\midrule
			+KW(4\texttimes) & 81.4 & 89.9 & 83.7 & 83.3 & 83.6 & 71.7 & 45.21M & +0.56G \\
			+FKW(4\texttimes) & \textbf{82.5} & \textbf{90.8} & \textbf{84.1} & \textbf{86.2} & \textbf{85.1} & \textbf{74.0} & 45.21M & +0.58G \\
			\bottomrule
		\end{tabular}
		
		\vspace{0.3cm}
		\footnotesize \textit{Note:} A "F" prefix before a dynamic convolution method indicates that the attention mechanism in that method has been replaced with FAT's frequency fusion approach.
		\label{table3}
	\end{table}
	
	\subsection{Experiment of Hi-CNA dataset}
	To further investigate model parameters and operational conditions, we conducted supplementary experiments on the Hi-CNA dataset\citep{sun2024}. Observations from the GID dataset\citep{Tong2020} revealed that optimal configurations of expert kernel counts and poolsize parameters vary across models. There is no doubt that when performance gains are similar, should give priority to the configurations with lower computational costs. We test our model on Hi-CNA, which has a smaller volume of data compared to the GID dataset.
	\\
	\\
	\textbf{CNN Backbones:}We compressed the ResNet18\citep{He2016} model to simulate resource-limited environments (e.g., mobile platforms). Specifically, we reduced the channel dimensions of the four ResNet layers in the encoder to [16, 32, 64, 128], with a correspondingly scaled-down U-Net\citep{Ronneberger2015} decoder. This lightweight architecture is denoted as ResNet18 (0.25\texttimes).
	\\
	\\
	\textbf{Experimental Setup:}In the encoder of our model, we replaced the ordinary convolutional kernels with FADConv and conducted experiments by setting the number of parallel expert convolutional kernels in FADConv to [2, 4, 6, 8] respectively. We selected the optimal number of expert kernels by balancing the computational cost and effect. Then, we set the poolsize to 8, 16, and 32 respectively for experiments to achieve a balance between computational cost and model performance. In the Hi-CNA dataset, the experiments were conducted under strictly controlled training settings: the basic learning rate was set to 0.0003, weight decay to 1e-4, batchsize to 16, and Epoch to 50 uniformly. No sample augmentation was used. The loss function adopted was BCE loss.
	\\
	\\
	\textbf{Results Comparison:}
	After scaling the width factor to 0.25x, the ResNet18+U-Net\citep{He2016,Ronneberger2015} architecture achieves a total parameter count of only 1.23M. We tested the model with varying numbers of expert kernels, and results (Table \ref{table4}) show that FADConv with 6 expert kernels delivers the best performance, improving F1-score and IoU by 1.62\% and 2.65\%, respectively. Notably, using 4 expert kernels incurs minimal performance degradation (0.02\% F1-score and 0.03\% IoU lower than 6 kernels) while reducing parameters by 1.17M, making it preferable for computation-constrained scenarios. To further investigate the impact of poolsize (p) in 2D DCT-based feature pooling, we fixed the expert kernel count to 4 and tested p values of [8, 16, 32]. As shown in Table \ref{table5}, p=16 achieves optimal performance while significantly reducing MAdds compared to p=32. Additionally, since ResNet-family models downsample 512×1512 images to a minimum feature size of 16×16, setting p=16 aligns with architectural constraints, ensuring computational efficiency.	
	\begin{table}[H]
		\centering
		\caption{A comparison of the results of ResNet18 models with different numbers of convolutional kernels based on FDConv. p is set to 16.
		All models are trained on the Hi-CNA dataset. Best results are bolded.}
		\footnotesize
		\begin{tabular}{llllllcl}
			\toprule
			\multicolumn{1}{l}{\textbf{Model}} & 
			\multicolumn{1}{c}{\textbf{Acc}} & 
			\multicolumn{1}{c}{\textbf{P}} & 
			\multicolumn{1}{c}{\textbf{R}} & 
			\multicolumn{1}{c}{\textbf{F1}} & 
			\multicolumn{1}{c}{\textbf{IoU}} & 
			\multicolumn{1}{c}{\textbf{Params}} & 
			\multicolumn{1}{l}{\textbf{MAdds}} \\
			\midrule
			\multicolumn{8}{l}{\textbf{ResNet18 (0.25x)}} \\
			Baseline & 92.51 & 88.17 & 88.67 & 88.42 & 79.24 & 1.23M & 5.97G \\
			+FDConv(2\texttimes) & 92.98 & \uline{90.10} & 87.89 & 88.98 & 80.15 & 1.85M & +8.02M \\
			+FDConv(4\texttimes) & \textbf{93.57} & 90.04 & \uline{90.01} & \uline{90.02} & \uline{81.86} & 3.01M & +9.18M \\
			+FDConv(6\texttimes) & \uline{93.54} & 89.42 & \textbf{90.67} & \textbf{90.04} & \textbf{81.89} & 4.18M & +10.35M \\
			+FDConv(8\texttimes) & 93.32 & \textbf{91.03} & 87.93 & 89.46 & 80.92 & 5.35M & +11.52M \\
			\bottomrule
		\end{tabular}
		\vspace{0.3cm}
		\label{table4}
	\end{table}
	\begin{table}[H]
		\centering
		\caption{The comparison of results when the expert convolution kernel number of FDConv is set to 4 and different pool sizes are used. All models are trained on the Hi-CNA dataset. Best results are bolded.}
		\footnotesize
		\begin{tabular}{llllllcl}
			\toprule
			\multicolumn{1}{l}{\textbf{Model}} & 
			\multicolumn{1}{c}{\textbf{Acc}} & 
			\multicolumn{1}{c}{\textbf{P}} & 
			\multicolumn{1}{c}{\textbf{R}} & 
			\multicolumn{1}{c}{\textbf{F1}} & 
			\multicolumn{1}{c}{\textbf{IoU}} & 
			\multicolumn{1}{c}{\textbf{Params}} & 
			\multicolumn{1}{l}{\textbf{MAdds}} \\
			\midrule
			\multicolumn{8}{l}{\textbf{ResNet18 (0.25x)}} \\
			Baseline & 92.51 & 88.17 & 88.67 & 88.42 & 79.24 & 1.23M & 5.97G \\
			+FDConv(4\texttimes, p=8) & \uline{93.36} & 89.25 & \textbf{90.28} & 89.76 & 81.42 & 3.01M & +3.22M \\
			+FDConv(4\texttimes, p=16) & \textbf{93.57} & \textbf{90.04} & \uline{90.01} & \textbf{90.02} & \textbf{81.86} & 3.01M & +9.18M \\
			+FDConv(4\texttimes, p=32)& 93.44 & \uline{89.68} & 89.98 & \uline{89.93} & \uline{81.54} & 3.01M & +56.89M \\
			\bottomrule
		\end{tabular}
		\vspace{0.3cm}
		\label{table5}
	\end{table}
	
	\subsection{Ablation Studies}
	Our previous experiments validated the effectiveness of FADConv and explored its underlying mechanisms, including parameter selection and computational efficiency trade-offs in lightweight models. To further verify the rationality of FADConv’s design, we conducted ablation experiments on the GID dataset, focusing on two key aspects:the effectiveness of frequency fusion strategies in the FAT module, and the superiority of FAT-based attention over traditional methods.
	\\
	\\
	\textbf{Experimental Setup:}We conducted experiments on the GID dataset using ResNet1-8+U-net\citep{He2016,Ronneberger2015} as the baseline. Performing 2D DCT on each channel yields a set of frequency feature vectors, which are compressed into a scalar representing the channel's global information. Three frequency fusion strategies were tested: direct summation, absolute summation, and fusion via a small network.
	\\
	\\
	\textbf{Results Comparison on frequency fusion:}
	As shown in Table \ref{table6}, the small network-based fusion achieves the best performance, outperforming direct summation by 1.2\% F1-score and 1.7\% IoU. We also observe that absolute summation surpasses direct summation, as it better represents the energy response intensity of frequency components by suppressing phase interference in the processed features. This validates the energy-centric perspective for aggregating frequency components. The small network balances contributions across all frequency bands, preventing over-reliance on the lowest-frequency components for energy representation. With negligible parameter and computational overhead, this method is selected as the frequency fusion strategy in the FAT module.
	\\
	\\
	\textbf{Results Comparison on attention:}
	Table \ref{table7} compares the performance of traditional GAP-based attention with FAT-based attention for dynamically combining expert convolution kernels. Using FAT improves the F1-score and IoU by 1.3\% and \\1.8\% over GAP. FAT captures richer global information through frequency analysis, enabling more rational weight allocation for expert kernels, which enhances feature extraction and generalization. Notably, even using the method of absolute summation(for example, when the effect of fusing frequency features using a small network is not satisfactory), it still won't perform worse than directly using GAP to obtain attention.
	\begin{table}[H]
		\centering
		\caption{The results obtained from experiments on the GID dataset by processing the frequency feature vectors using the FCANet frequency fusion method, direct frequency addition, addition of frequency absolute values, and small network fusion methods respectively for comparison. Best results are bolded.}
		\footnotesize
		\begin{tabular}{lllllll}
			\toprule
			\multicolumn{1}{l}{\textbf{Model}} & 
			\multicolumn{1}{c}{\textbf{OA}} & 
			\multicolumn{1}{c}{\textbf{Acc}} & 
			\multicolumn{1}{c}{\textbf{P}} & 
			\multicolumn{1}{c}{\textbf{R}} & 
			\multicolumn{1}{c}{\textbf{F1}} & 
			\multicolumn{1}{c}{\textbf{IoU}} \\
			\midrule
			Baseline & 80.9 & 89.5 & 81.3 & 85.5 & 83.3 & 71.5 \\
			+FDConv (FCA) & 81.3 & 89.8 & 81.3 & 86.5 & 83.8 & 72.2 \\
			+FDConv (add) & 81.5 & 90.1 & 82.7 & 85.6 & 84.1 & 72.6 \\
			+FDConv (add of abs) & 81.8 & 90.3 & 83.4 & 85.4 & 84.4 & 73.0 \\
			+FDConv (fusion) & \textbf{82.5} & \textbf{90.9} & \textbf{84.4} & \textbf{86.1} & \textbf{85.3} & \textbf{74.3} \\
			\bottomrule
		\end{tabular}
		
		\vspace{0.3cm}
		\footnotesize \textit{Note:} FCANet pioneered the incorporation of frequency domain features into channel attention mechanisms. Thus We compare the effectiveness of its attention generation approach.
		\label{table6}
	\end{table}

	\begin{table}[H]
		\centering
		\caption{The effectiveness of each module was tested by gradually adding dynamic convolution and our FAT module to ResNet18. The experimental results on the GID dataset are presented. Best results are bolded.}
		\footnotesize
		\begin{tabular}{lllllll}
			\toprule
			\multicolumn{1}{l}{\textbf{Model}} & 
			\multicolumn{1}{c}{\textbf{OA}} & 
			\multicolumn{1}{c}{\textbf{Acc}} & 
			\multicolumn{1}{c}{\textbf{P}} & 
			\multicolumn{1}{c}{\textbf{R}} & 
			\multicolumn{1}{c}{\textbf{F1}} & 
			\multicolumn{1}{c}{\textbf{IoU}} \\
			\midrule
			Baseline & 80.9 & 89.5 & 81.3 & 85.5 & 83.3 & 71.5 \\
			Baseline + dynamic conv & 81.6 & 90.1 & 83.0 & 85.0 & 84.0 & 72.5 \\
			Baseline + FAT + dynamic conv & \textbf{82.5} & \textbf{90.9} & \textbf{84.4} & \textbf{86.1} & \textbf{85.3} & \textbf{74.3} \\
			\bottomrule
		\end{tabular}
		\vspace{0.3cm}
		\label{table7}
	\end{table}
	\section{Conclusion}
	In this paper, we propose Frequency-Aware Dynamic Convolution (FADConv), a novel dynamic convolution architecture with a frequency-guided attention mechanism, to enhance the performance of deep CNNs. FADConv leverages 2D DCT to extract frequency features from input tensors, enabling high-quality global representation and optimized attention weight allocation for dynamic kernels. It achieves significant improvements while maintaining minimal computational overhead and can be seamlessly integrated into existing CNNs by replacing standard convolutions. Notably, integrating our frequency-aware attention module into other dynamic convolution frameworks also boosts their performance. Experiments on the GID dataset and Hi-CNA dataset demonstrate its potential: FADConv substantially improves cropland extraction accuracy compared to prior dynamic convolution methods, effectively addressing cropland non-agriculturalization challenges. We hope this work will offer insights for future research on dynamic convolutions and contribute to practical solutions for cropland non-agriculturalization.
	
	\section*{Acknowledgements}
	This work was supported by the SRTP project of Southwest Jiaotong University.We thank the research team of Sun et al. for granting us permission to utilize the Hi-CNA dataset in our semantic segmentation research. Their support has been instrumental in enabling the experimental validation of our methodology.

\end{document}